\documentclass[runningheads]{llncs}
\usepackage{graphicx}
\usepackage{comment}
\usepackage{amsmath}
\usepackage{booktabs}
\usepackage{multirow}
\usepackage{amssymb}
\begin{document}
\title{GMFVAD: Using Grained Multi-modal Feature to Improve Video Anomaly Detection}
\titlerunning{GMFVAD: Grained Multi-modal Feature in VAD}
%
\author{Guangyu Dai\inst{1} \and
Dong Chen\inst{1,2} \and
Siliang Tang\inst{1}\and Yueting Zhuang\inst{1}}
\authorrunning{G.Dai et al.}
%
\institute{Zhejiang University, China\and Zhengzhou University, China \\
\email{\{daiguangyu,chendongcs,siliang,yzhuang\}@zju.edu.cn}}
\maketitle              
\begin{abstract}
Video anomaly detection (VAD) is a challenging task that detects anomalous frames in continuous surveillance videos. Most previous work utilizes the spatio-temporal correlation of visual features to distinguish whether there are abnormalities in video snippets. Recently, some works attempt to introduce multi-modal information, like text feature, to enhance the results of video anomaly detection. However, these works merely incorporate text features into video snippets in a coarse manner, overlooking the significant amount of redundant information that may exist within the video snippets.  Therefore, we propose to leverage the diversity among multi-modal information to further refine the extracted features, reducing the redundancy in visual features, and we propose Grained Multi-modal Feature for Video Anomaly Detection (GMFVAD).  Specifically, we generate more grained multi-modal feature based on the video snippet, which summarizes the main content, and text features based on the captions of original video will be introduced to further enhance the visual features of highlighted portions. Experiments show that the proposed GMFVAD achieves state-of-the-art performance on four mainly datasets. Ablation experiments also validate that the improvement of GMFVAD is due to the reduction of redundant information.

\keywords{video anomaly detection\and grained multi-modal feature \and weakly supervised learning}
\end{abstract}
\section{Introduction}
Video anomaly detection(VAD) in surveillance videos plays a crucial role in various fields, such as manufacturing and security. For example, in terms of security, VAD can help people respond more quickly to abnormal events, such as violence and crime, and enables the rapid dispatch of security personnel to mitigate potential losses. \\ 
VAD is a challenging task that detects few anomalous frames in continuous surveillance videos. Due to the high cost of frame-level annotation, the prevailing approach in supervised VAD relies on weakly supervised settings with video-level annotation. Previous works commonly extract spatio-temporal visual features from videos and enhance anomaly detection performance by considering the spatio-temporal correlation \cite{tian2021weakly}, or by analyzing global and local scenes \cite{chen2023mgfn}. Recently, TEVAD \cite{chen2023tevad} proposes to incorporate text features to enhance VAD performance, as texts are semantically rich, while visual features are unable to capture semantic meanings.\\
However, TEVAD merely incorporates text features into video snippets in a coarse manner, overlooking the significant amount of redundant information in the video snippets. Such redundant information makes it challenging for the summarized text features to align with all visual features of video snippets. On the contrary, the diversity among multi-modal feature also makes it possible for attenuating visual redundant information with the help of text. Based on previous works that distinguish anomalies with the spatio-temporal features and multi-modal information, we propose Grained Multi-modal Feature for Video Anomaly Detection (GMFVAD) to further enhance the performance of VAD by attenuating visual redundant information with text features.\\
Specifically, we first employ the glance-focus network to scan and locate video snippets, identifying snippets that are more likely to contain anomalies as a visual feature. Then, we generate dense captions with Swinbert \cite{lin2022swinbert}, and fuse multi-modal features to weakening the redundant visual features. Thus, the multi-modal features will better focus on important information to distinguish normal and abnormal events.The contributions of our work are outlined as: 
\begin{itemize}
    \item We propose GMFVAD, a weakly supervised framework for video anomaly detection. Different from prior works, GMFVAD enhances the performance of VAD with grained multi-modal feature. 
    \item We use both visual and text feature in our GMFVAD network. When we generating the visual feature in GMFVAD model, we implement glance-focus network to generate more grained visual feature. We use SwinBert to generate text feature, both method are proved significant in our experiments.
    \item The proposed method achieves state-of-the-art performance on various datasets. In addition, our ablation experiment demonstrates the efficacy of incorporating both visual and text features in enhancing the performance of VAD, and GMFVAD proves to be more effective than only using single modal feature.
\end{itemize}

\section{Related Works}
Prior to the advent of deep learning, VAD was regarded as a single-class classification problem relying on manual features\cite{basharat2008learning,medioni2001event}. Nowadays, most studies uses deep learning to solve VAD problem, which can be categorized into unsupervised methods and weakly-supervised methods based on whether there are video-level labels.
\subsection{Unsupervised VAD Methods}
The majority of unsupervised methods for VAD are built upon video reconstruction or future frame prediction \cite{Burlina_2019_CVPR,zong2018deep,gong2019memorizing,ionescu2019object,liu2018future,nguyen2019anomaly,nguyen2019anomaly2}. Typically, videos in training set is encoded by autoencoder to obtain their representations, and the representations are utilized for video reconstruction or frame prediction. 
During inference, if the video is abnormal, the corresponding reconstruction error will be high, and vice versa.
Reconstruction-based methods assume that normal videos follow the distributions of the training data, while abnormal videos do not follow such distributions.
However, this assumption is not always correct, as autoencoder may overfit on some distributions that are the most prevalent in the training data. To alleviate such issue, other unsupervised methods for VAD are proposed. \cite{wu2020not,astrid2021learning,georgescu2021background,zaheer2020old} proposes to generate pseudo labels and improve VAD by pseudo-supervised training. Additionally, \cite{abati2019latent,tian2020few,bergmann2019mvtec} imposes constraints on the latent space of the normal manifold to acquire compact representations of normal data.
\subsection{Weakly-Supervised VAD Methods}
As unsupervised methods are unable to effectively capture the characteristics of abnormal distributions, some studies turn to focus on supervised ways. Considering the high cost of frame-level label annotation, researchers annotate videos with video-level labels, which is weakly supervision. Zhong et al.\cite{zhong2019graph} implement graph convolution network (GCN) for noise removal. However, GCN introduces more expensive computational costs and overfitting issues. 
Sultani et al.\cite{sultani2018real} and Wu et al.\cite{wu2020not,wu2022weakly} proposed multi-task model to solve the problem. More  further works\cite{georgescu2021anomaly,chang2021contrastive,lv2021localizing,wu2021learning,sapkota2022bayesian,panariello2022consistency} are proposed based on the multi-task method.
Another studies\cite{tian2021weakly,chen2023mgfn,li2022self,zhang2019temporal} proposed multiple instance learning (MIL) framework to deal with weakly supervised VAD task. The methods learned spatio-temporal features in the video as auxiliary information to enhance the performance of VAD. 
Recently, some studies try to implement cross-modal information to improve VAD performance. Based on previous works, Chen et al.\cite{chen2023tevad}, Yuan et al.\cite{yuan2024surveillance} propose to incorporate text features to enhance the semantic information. Acsintoae et al. \cite{acsintoae2022ubnormal} proposes UBnormal dataset and new open-set VAD task as a new benchmark. With the development of large-scale model pretraining, some works\cite{joo2023clip,wu2024vadclip,Wu_2024_CVPR,lv2023unbiased} tried to utilize the powerful visual feature from CLIP\cite{radford2021learning} to improve VAD performance in their studies, and \cite{tang2024hawk,zhang2024holmes,wu2024weakly,yuan2024surveillance} implement methods based on multi-modal LLM.
\section{Method}
\begin{figure}[htbp]
  \centering
  \includegraphics[width=\textwidth]{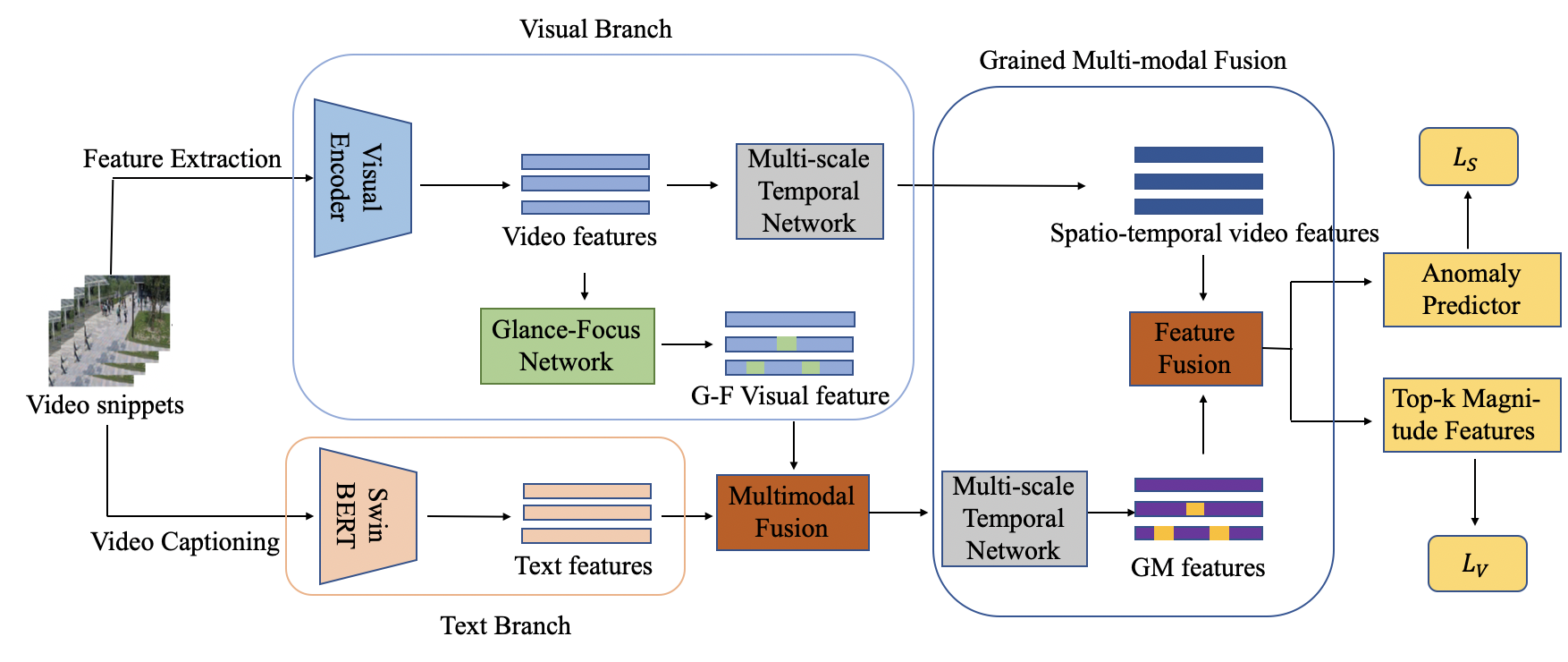}
  \caption{The overview of our proposed GMFVAD method. We input the video feature and text feature to generate grained multi-modal feature ,use MTN to fuse grained multi-modal feature and video feature, and implement top-k magnitude features to calculate loss.}
  \label{fig:fig1}
\end{figure}
Our proposed Grained Multi-modal Feature for Video Anomaly Detection (GMFVAD) is presented in Fig.1. Based on the multi-modal methods, we further discuss how to use text information summarized from video snippets to mitigate the impact of visual redundancy on anomaly detection. $V=\{(\boldsymbol{F_i},y_i)\}_{i=1}^{|V|}$ denotes training data of a video, where ${\boldsymbol{F}}$ denotes features extracted by pre-trained I3D\cite{carreira2017quo}, and $y$ denotes binary video-level label. We input $\boldsymbol{F}$ into grained multi-modal feature generation part to obtain grained multi-modal feature $\boldsymbol{F}_{GM}$  Subsequently, we input both $\boldsymbol{F}$ and $\boldsymbol{F}_{GM}$ into multi-scale temporal network (MTN) to capture multi-scale temporal dependencies and fuse output features as $X$.
Finally, we select the top-k largest magnitude feature in $X$ to train a classifier and calculate loss function.
\subsection{Video Feature Generating}
During step of extracting the visual feature of video, we implement resnet50\cite{he2016deep} backbone to extract I3D features. Consistent with previous effective VAD methods \cite{tian2021weakly,chen2023tevad,feng2021mist}, we apply ten-crop augmentation on the dataset to enhance model performance. Firstly, we crop the four corners and the central of the frame, resulting in five-crop augmentation. And then we include the horizontally flipped version of the five-crop to obtain ten-crop augmentation.\\
Other extractors such as C3D\cite{tran2015learning}, TSN\cite{wang2018temporal} also can be implemented in visual feature extracting part, however, the whole model performed better when we implement I3D feature. Therefore, we use I3D as our default experiment configuration.
\subsection{Grained Visual Feature Generation} 
In visual modal, we implement glance-focus network to generate visual part of grained multi-modal feature.$\boldsymbol{F}_{GF}=f_{GF}(\boldsymbol{F})$ refers to the feature output from glance-focus network when we set $\boldsymbol{F}$ as input. Glance-Focus Network is proposed in \cite{chen2023mgfn} to improve the performance of VAD task. The brief architecture of glance-focus network is presented in Fig.2. 
\begin{figure}[htbp]
  \centering
  \includegraphics[width=\textwidth]{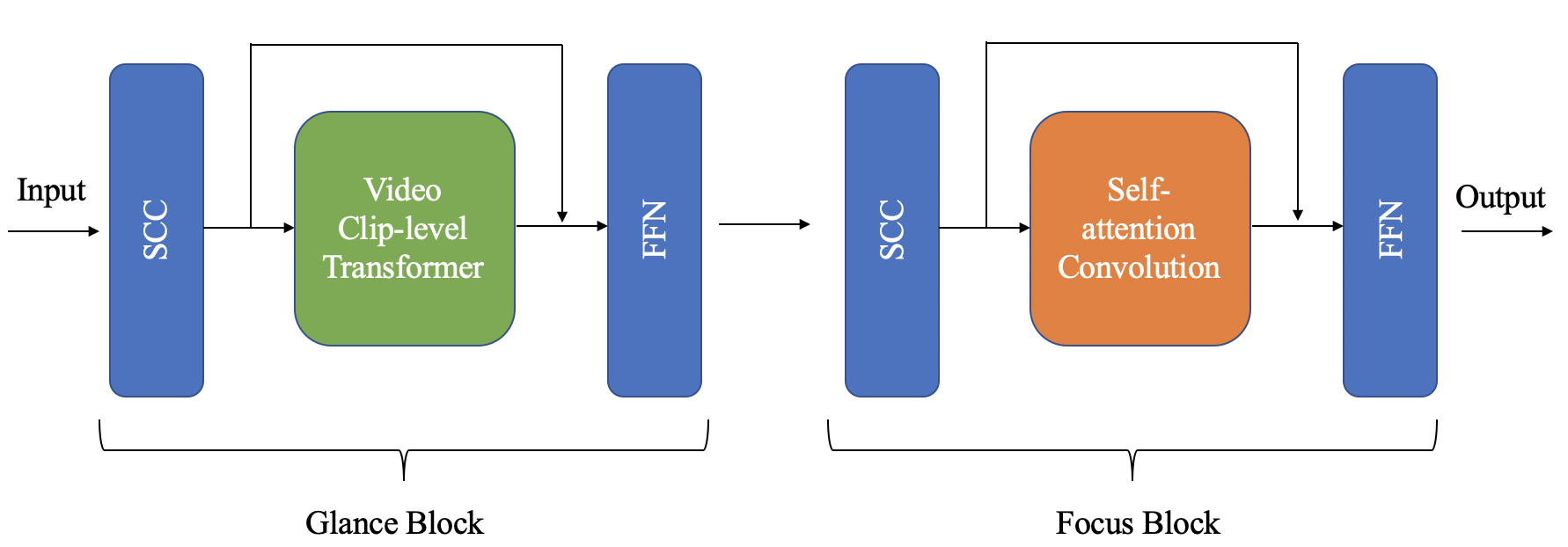}
  \caption{The architecture of glance-focus network. It consists a video transformer(glance block) and a self-attention convolution network(focus block), and using FFN and SCC to align them.}
  \label{fig:fig2}
\end{figure}
Glance-focus network consists of two parts, glance block and focus block.Glance block consists of a short-cut convolution(SCC), a video clip-level transformer, an additional Feed-Forward Network (FFN) including two fully-connected layers and a GeLU non-linear function to further improve the model's representation capability. The output feature $f_{G}(\boldsymbol{F})$is fed to the following Focus block.Focus block consists of a short-cut convolution (SCC), a self-attentional convolution (SAC), and a FFN. With $f_{G}(\boldsymbol{F})$ as input, the focus block first increase the channel number to C/16 with a convolution, and SCC generates the temporary feature $f_{SCC}(\boldsymbol{F})$.After that, the focus block implements self-attention convolution to enhance the feature of each clip.The self-attention convolution allows each clip to get access to the nearby clip so that the correlation between different clips can be learned. After a 2-layer FFN, the block outputs the glance-focus feature $f_{GF}(\boldsymbol{F})$. \\
With the help of glance-focus network, the output feature $\boldsymbol{F}_{GF}=f_{GF}(\boldsymbol{F})$ is more grained. Glance block provides the network with the knowledge of “what the normal cases are like” to better detect the abnormal events, and focus block combines one video clip and other nearby clips in self-attention, which highlights the correlations between different clips.
\subsection{Text Feature Generating}
In our text feature generating part, we implement SwinBERT\cite{lin2022swinbert} pre-trained on VATEX\cite{wang2019vatex}, which is a large-scale and general video dataset and provides general video captioning capacity, to generate dense video captioning. Then we use SimCSE, a contrastive learning method to generate sentence embedding set, the output sentence embedding is the text feature of GMFVAD model as $\boldsymbol{F}_{txt}$. 
\subsection{Grained Multi-modal Feature Fusing}
After extracting grained visual feature of video from glance-focus network and text feature from SwinBERT, we employ the late fusion scheme\cite{bakkali2020visual} to fuse the features together as grained multi-modal  feature, and then use multi-scale temporal feature learning (MTN) to fuse video feature and grained multi-modal feature. To align with the five/ten cropped visual features, the text features are also tiled for five/ten times. We concatenate visual feature and text feature as:\\
\begin{equation}
    \boldsymbol{F}_{GM} = \{\boldsymbol{F}_{GF}|\boldsymbol{F}_{txt}\}
\end{equation}
\par
\subsection{Multi-modal Multi-Scale Temporal Feature Learning} 
Our work implements multi-scale temporal network(MTN) to process the visual features of the video and grained multi-modal features. Fig.3 shows a brief structure of MTN.
\cite{tian2021weakly} introduces MTN for the first time and provides theoretical evidence of the significance and effectiveness of capturing multi-scale temporal dependencies among adjacent video clips for anomaly detection.
\begin{figure}[htbp]
  \centering
  \includegraphics[width=\textwidth]{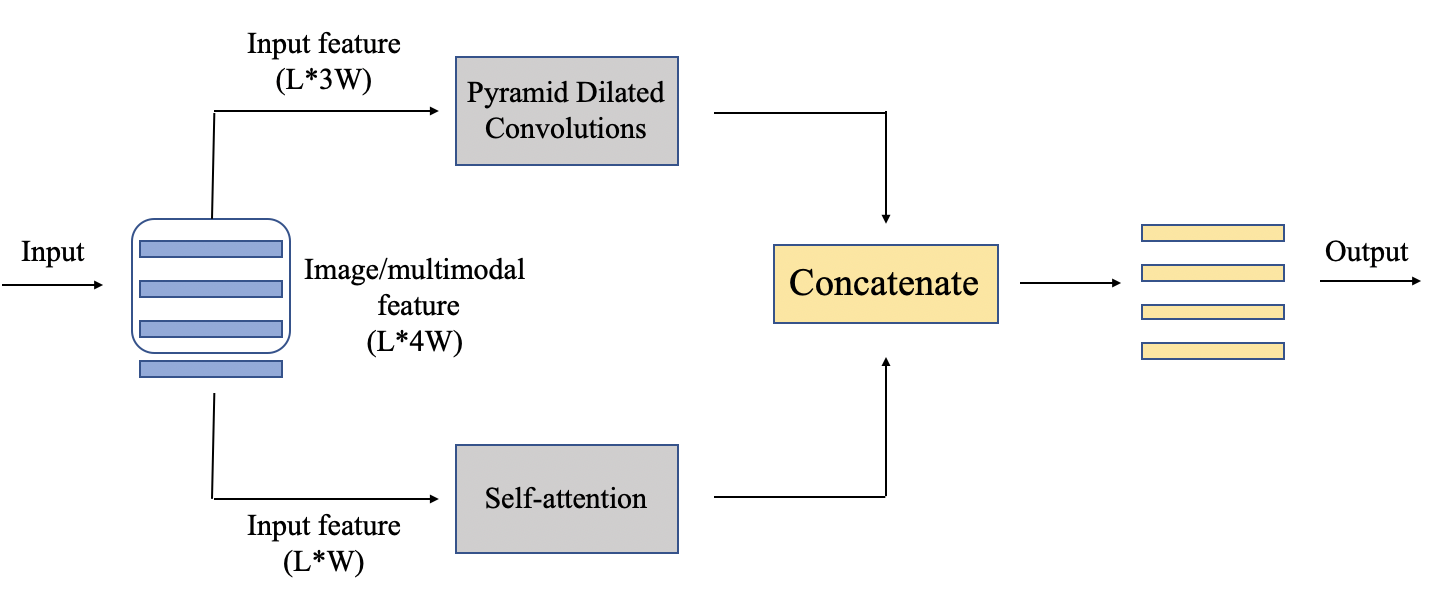}
  \caption{A brief architecture of MTN.The input feature is divided into two parts,   three quarter of them are input to pyramid dilated convolution part, and the last one quarter are input to self-attention part.}
  \label{fig:fig3}
\end{figure}
The visual MTN is composed of two parts. The first part utilizes a 3-layer pyramid dilated convolutions to extract temporal dependencies between different snippets. This aims to capture the temporal relationships at various scales. The other part incorporates a non-local block layer that calculates global time correlation using a self-attention network, the output dimension of each layer is a quarter of the original feature, we concatenate the output of each layer and obtain the MTN feature $X_v=f_{MTN}(\boldsymbol{F})$ of the video.
In order to capture the short-term and long-term dependencies between feature snippets of the multi-modal feature, We implement a similar MTN network to process the grained multi-modal feature.We set $X_{GM}=f_{MTN}(\boldsymbol{F}_{GM})$ as the output when inputting $\boldsymbol{F}_{GM}$,then fuse the video output feature and grained multi-modal output feature using a fully-connect layer:
\begin{equation}
    X = FC(X_{GM};\theta) + X_v
\end{equation}
$FC$ denotes a fully-connect layer with parameter $\theta$. 
\subsection{Loss Function}
Following previous work\cite{tian2021weakly}, We choose top-k magnitude feature snippets as set $X_{
topk}$ to calculate our loss function. Our loss function consists of magnitude loss between normal video and abnormal video, and classifier loss between snippets.The feature magnitude of video feature V is computed as:
\begin{equation}
    m_k(X) = \frac{1}{k}\sum_{x_n\in X_{topk}}||x_n||^2
\end{equation}
The loss function of our model consists of two components: $L_{v}$ and $L_{s}$. $L_{v}$  the sum of magnitude difference between normal and abnormal videos$l_v$, calculated as :
\begin{equation}
    l_{v} = 
\begin{cases}
max(0,c-(m_k(X_i)-m_k(X_j))),& y_i=1, y_j=0 \\
0,& \text{otherwise}
\end{cases}
\end{equation}
\begin{equation}
    L_{v} = \sum_{i,j=1}^{|V|}l_{v}(X_i,X_j,y_i,y_j)
\end{equation}
$L_{s}$ is based on the cross-entropy loss in the anomaly detection classifier of the top-k snippets. 
\begin{equation}
    L_{s}=\sum_{x_n\in X_{topk}}-(ylog(f_c(x_n))+(1-y)log(1-f_c(x_n)))
\end{equation}
$f_c$ is a 3-layer fully-connect network as classifier. The loss function is defined as follow:
\begin{equation}
    L = \alpha L_{v} + L_{s}
\end{equation}
where $\alpha$ is variable to balance the loss terms.
\section{Experiments}
\label{sec:experiments}

\subsection{Benchmark Datasets}
We validate the effectiveness of the proposed method on four datasets, UCSD-Peds, ShanghaiTech, UCF-Crime and XD-Violence.\\
\subsubsection{UCSD-Peds}
UCSD-Peds is a small dataset for VAD task, it comprises two subsets: UCSD-Ped1 and UCSD-Ped2. UCSD-Ped1 consists of 70 videos, while UCSD-Ped2 consists of 28 videos. In our experiment, we randomly allocate 6 abnormal videos and 4 normal videos from the UCSD-Ped2 dataset as the training set, the remaining videos as the test set.\par
\subsubsection{ShanghaiTech}
ShanghaiTech is a medium-scale VAD dataset designed for unsupervised VAD setting, where the videos come from street video surveillance. This dataset consists 307 normal videos and 130 abnormal videos.For weakly-supervised setting, we follow \cite{zhong2019graph} ,reorganize the dataset by selecting a subset of anomalous testing videos into training data to build a weakly supervised training set. Specifically, we divide the dataset into a training set of 238 videos and a test set of 199 videos.\par
\subsubsection{UCF-Crime}
UCF-Crime is a large-scale VAD dataset,containing 1900 videos with a total duration of 128 hours. The video data in UCF-Crime is sourced from surveillance videos, similar to ShanghaiTech dataset. 
The UCF-Crime dataset contains a training set of 1610 videos and a test set of 290 videos, with a total of 13 crime-related abnormal events. It is worth noting that in UCF-Crime dataset, the training set has video-level labels, and the test set has frame-level labels.Thus, we can obtain frame-level AUC results in the experiment conducted on the UCF-Crime dataset.\par
\subsubsection{XD-Violence}
XD-Violence is a comprehensive VAD dataset, featuring a large-scale collection of diverse scenarios. It contains real-life movies from online videos, sports streams, surveillance cameras and CCTVs, and the dataset covers 6 types of abnormal events related to violence. The dataset consists of 4754 videos, totaling over 217 hours in duration. The training set consists of 3754 videos with video-level labels, while the test set comprises 800 videos with frame-level labels. Additionally, XD-Violence is a dataset including both video and audio modal data. In our experiments, we only use the video data and compare our results exclusively with methods that solely use video data.
\subsection{Evaluation Measures}
We compare our method and baselines by the Area Under the Curve (AUC) metric. Additionally, follow prior works \cite{tian2021weakly,chen2023mgfn,sultani2018real}, we employ Average Precision (AP) as the evaluation metric for XD-Violence. In the context of VAD, higher AUC and AP values indicate better performance.

\subsection{Implementation Details}
We use pytorch \cite{paszke2019pytorch} 
to train our model on a single 2080ti GPU, and we use Adam with a batch size of 64, learning rate of 0.001, and weight decay of 0.005 to optimize our model. \\
When extracting visual feature, we split the video into T snippets, and a snippet has 16 frames. For UCSD-Ped2, ShanghaiTech and UCF-Crime dataset,we use I3D feature extractor with ResNet50 backbone pretrained on Kinetic-400\cite{kay2017kinetics}; For XD-Violence dataset, we use the I3D features provided by the author. When We generate dense caption, we implement default setting for SwimBert on VATEX, and we use default setting of supervised SimCSE in sentence embedding generation. We set dilation parameter in MTN as 1,2,4 respectively, and set $\alpha$=0.0001 in loss function.
\subsection{Results on benchmark datasets}
Table 1 presents the AUC result comparison on UCSD-Ped2, Shanghaitech, UCF-Crime dataset, and Table 2 shows the AP experiment result comparison on XD-Violence dataset. Subsequently, we will conduct a detailed analysis of the results obtained from each individual dataset.\\
\begin{table*}[!t]
  \centering
  \begin{tabular}{cc|ccc}
    \toprule
    \multirow{2}{*}{Supervision}   & \multirow{2}{*}{Method}   & \multicolumn{3}{c}{AUC(\%)} \\
    \multirow{2}{*}{}   & \multirow{2}{*}{} & UCSD-Ped2 & ShanghaiTech & UCF-Crime   \\
    \midrule
    \multirow{2}{*}{Unsupervised} &  {SSMTL}$^*$\cite{georgescu2021anomaly} & 97.52 & 82.45 & - \\
    \multirow{2}{*}{} & {Georgescu et al.}$^*$\cite{georgescu2021background}       & 98.70  & 83.54 & -  \\
    \midrule
    \multirow{6}{*}{Weakly-supervised} & GCN-Anomaly\cite{zhong2019graph} & 93.22 & 84.61 & 75.25 \\
    \multirow{6}{*}{} & Sultani et al. \cite{sultani2018real}& 92.28 & 85.52 & 75.41  \\
    \multirow{6}{*}{} & MIST\cite{feng2021mist} & - & 93.38 & 77.25  \\
    \multirow{6}{*}{} & RTFM\cite{tian2021weakly}&98.60 & 97.21 & 84.30 \\
    \multirow{6}{*}{} & MGFN\cite{chen2023mgfn} & - & - & 84.47 \\
    \multirow{6}{*}{} & TEVAD\cite{chen2023tevad} & 98.70 & 98.10 & 84.90 \\
    \multirow{6}{*}{} & DMU\cite{zhou2023dual} & - & - & 85.14 \\
    \multirow{6}{*}{} & OVVAD\cite{Wu_2024_CVPR} & - & - & 86.40 \\
    \multirow{6}{*}{} & GMFVAD(Ours) & \textbf{98.85} & \textbf{98.23} &\textbf{87.22}  \\
    \bottomrule
  \end{tabular}
  \label{tab:table1}
  \caption{AUC result comparison on UCSD, Shanghai, UCF-Crime dataset.}
\end{table*}
\begin{table}
  \caption{AP Result comparison on XD-Violence dataset.}
  \centering
  \begin{tabular}{lll}
    \toprule
    Supervision     & Method   & AP ($\%$) \\
    \midrule
    \multirow{6}{*}{weakly-supervised} & Wu et al \cite{wu2021learning}& 75.41\\ 
    \multirow{6}{*}{} & Sultani et al. \cite{sultani2018real}& 75.68\\
    \multirow{6}{*}{} & RTFM \cite{tian2021weakly}& 77.81 \\
    \multirow{6}{*}{} & TEVAD\cite{chen2023tevad} & 79.80\\
    \multirow{6}{*}{} & MGFN\cite{chen2023mgfn} & 80.11 \\
    \multirow{6}{*}{} & DMU\cite{zhou2023dual} & \textbf{81.66} \\
    \multirow{6}{*}{} & GMFVAD(Ours) & 81.52  \\
    \bottomrule
  \end{tabular}
\end{table}
\\
\textbf{UCSD-Ped2 dataset.} Due to the dataset’s early stage and small-scale, most methods can achieve over 90\% results on this dataset, thereby limiting its ability to effectively evaluate the models.However, GMFVAD  performs a slight advantage over the best existing unsupervised and weakly-supervised methods. \\
\textbf{ShanghaiTech dataset.} Compared with UCSD-Ped2, ShanghaiTech demonstrates greater capability to evaluate model effects, and weakly-supervised methods clearly outperforming unsupervised methods.Our method achieves best result in the comparison of unsupervised and weakly-supervised methods.\\
\textbf{UCF-Crime dataset.} Due to all the weakly-supervised method outperforming unsupervised method 5\% or more, we exclusively present the results of weakly-supervised method in the table. Our method achieves state-of-the-art results in this complex video anomaly detection dataset.\\
\textbf{XD-Violence dataset.} Our approach outperforms the SOTA methods that solely rely on visual features, resulting in a 1.41\% AP improvement, this result demonstrates the effectiveness of our multi-modal feature. However. our method slightly lower than multi-modal method DMU\cite{zhou2023dual}. 

\begin{figure}[htbp]
  \includegraphics[width=\textwidth]{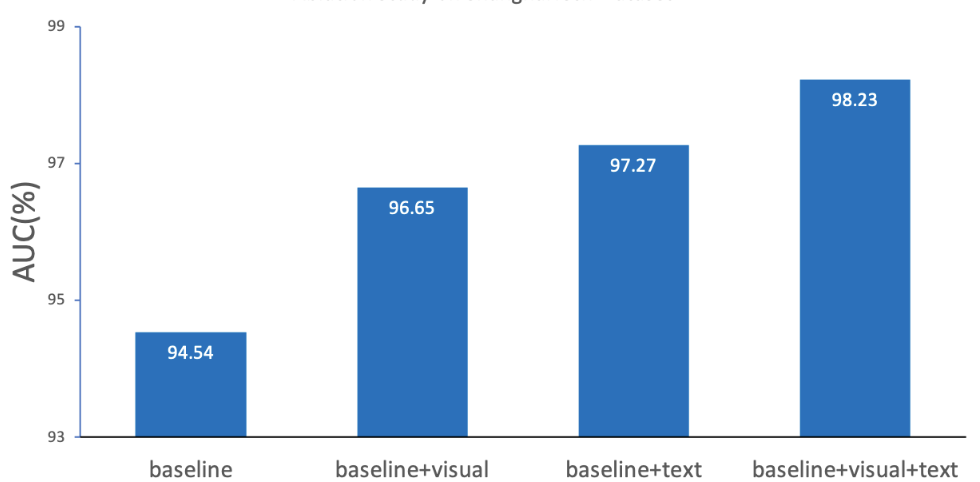}
  \caption{The ablation study result on ShanghaiTech dataset.}
  \label{fig:fig4}
\end{figure}
\begin{figure}[htbp]
  \includegraphics[width=\textwidth]{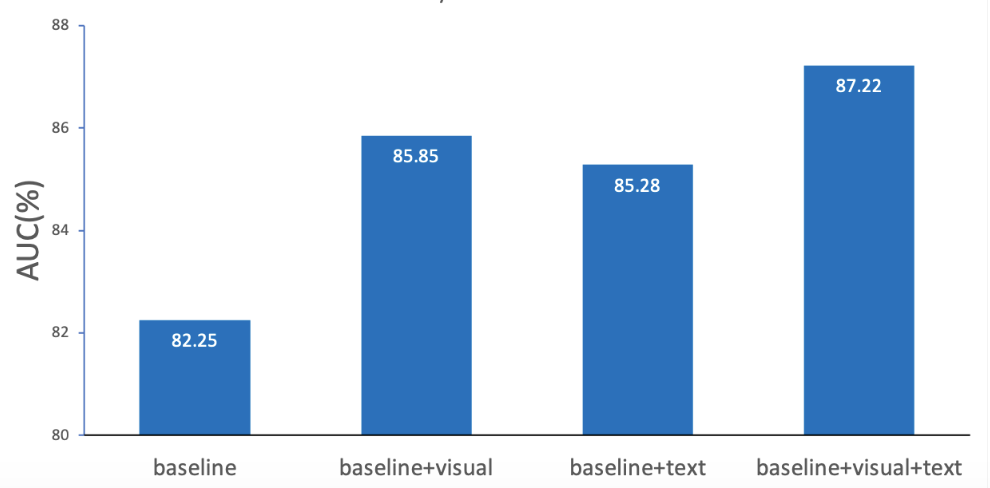}
  \caption{The ablation study result on UCF-Crime dataset.}
  \label{fig:fig5}
\end{figure}
\subsection{Ablation Experiment}
To demonstrate the roles of visual feature and text feature in GMFVAD, we conduct ablation experiments on ShanghaiTech and UCF-Crime datasets. The results are presented on Fig.4 and Fig.5 respectively. It can be seen that incorporating visual feature and text feature can significantly outperforms the baseline, by 3.59\% and 4.97\%, on ShanghaiTech and UCF-Crime respectively.
The experimental results clearly demonstrate that when incorporating both visual and textual modalities to reduce redundancy leads to better performance in video anomaly detection, surpassing the method of not use or use solely a single redundancy reduction method.Our ablation experiment illustrates that the visual glance-focus network and text captions are capable of effectively reducing redundant information within visual features, resulting in improved accuracy for anomaly detection.\\
Additionally, the text feature contributes more on the AUC increase of the ShanghaiTech, whereas the visual feature plays a more significant role in the AUC increase of the UCF-Crime.Based on this phenomenon, we conjecture that for datasets with relatively simple visual feature, text information exhibit more pronounced influence, and for datasets with complex visual information, visual part plays more significant role than text part in multi-modal feature.

\section{Conclusion}
Video Anomaly Detection(VAD) is a challenging task with a wide range of real-life applications. Previous works focused more in visual feature and overlooked information hided in text. Some works that consider multi-modal information for VAD always overlook that the redundant information in video snippets may influence the performance of VAD model negatively. \\
In this work, we propose a weakly supervised anomaly detection framework, GMFVAD, which leverages the diversity among multi-modal information to further refine the extracted features and enhance the performance of VAD.GMFVAD combines visual and text feature as a multi-modal feature, combined with the MTN architecture to better utilize multi-modal features for video anomaly detection. GMFVAD implements the glance-focus network to enhance the quality of text features using visual information. This approach enables a more fine-grained analysis, emphasizing text information that is more related to abnormal part of full video. Finally, GMFVAD generate model loss through top-k MIL framework. \\
As a result, GMFVAD achieves improved performance in completing VAD tasks by leveraging multi-modal feature.We evaluate GMFVAD on four main VAD datasets and the proposed GMFVAD method achieves state-of-the-art performance in majority cases.
%
%
%
\bibliographystyle{splncs04}
\bibliography{refs}
%
\end{document}